\documentclass[10pt,twocolumn,letterpaper]{article}

\usepackage[pagenumbers]{iccv} 

\definecolor{iccvblue}{rgb}{0.21,0.49,0.74}
\usepackage[pagebackref,breaklinks,colorlinks,allcolors=iccvblue]{hyperref}

\usepackage{multirow}
\usepackage[normalem]{ulem}
\useunder{\uline}{\ul}{}
\usepackage{booktabs}
\usepackage{url}
\usepackage{amsmath}
\usepackage{utfsym}
\usepackage{pifont}

\def\paperID{7556} 
\def\confName{ICCV}
\def\confYear{2025}

\title{CoTMR: Chain-of-Thought Multi-Scale Reasoning for Training-Free \\ Zero-Shot Composed Image Retrieval}

\author{Zelong Sun\thanks{Equal Contribution} ,   Dong Jing$^{\ast}$,    Zhiwu Lu \thanks{Corresponding Author} \\
Gaoling School of Artificial Intelligence \\
Renmin University of China, Beijing, China\\
{\tt\small zelongsun@ruc.edu.com, luzhiwu@ruc.edu.com}}

\begin{document}
\maketitle
\begin{abstract}
Zero-Shot Composed Image Retrieval (ZS-CIR) aims to retrieve target images by integrating information from a composed query (reference image and modification text) without training samples. 
Existing methods primarily combine caption models and large language models (LLMs) to generate target captions based on composed queries but face various issues such as incompatibility, visual information loss, and insufficient reasoning. 
In this work, we propose CoTMR, a training-free framework crafted for ZS-CIR with novel Chain-of-thought (CoT) and Multi-scale Reasoning.
Instead of relying on caption models for modality transformation, CoTMR employs the Large Vision-Language Model (LVLM) to achieve unified understanding and reasoning for composed queries. 
To enhance the reasoning reliability, we devise CIRCoT, which guides the LVLM through a step-by-step inference process using predefined subtasks.
Considering that existing approaches focus solely on global-level reasoning, our CoTMR incorporates multi-scale reasoning to achieve more comprehensive inference via fine-grained predictions about the presence or absence of key elements at the object scale.
Further, we design a Multi-Grained Scoring (MGS) mechanism, which integrates CLIP similarity scores of the above reasoning outputs with candidate images to realize precise retrieval. 
Extensive experiments demonstrate that our CoTMR not only drastically outperforms previous methods across three prominent benchmarks but also offers appealing interpretability.
\end{abstract}    

\begin{figure}[t!]
\centering  
\includegraphics[width=0.99\linewidth]{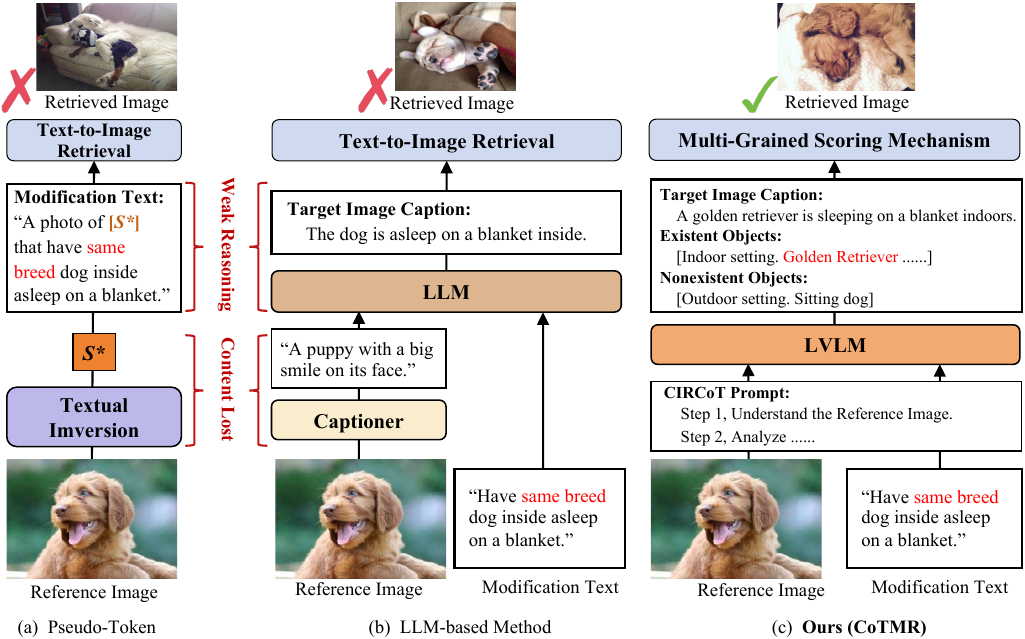} 
\vspace{-0.08in}
\caption{\textbf{Flowcharts of existing ZS-CIR methods and our proposed CoTMR.} Methods (a) and (b) face serious issues of visual information loss and insufficient reasoning. In contrast, our method (c) fully perceives image content, enhances reasoning process with CIRCoT, and augments multi-grained descriptions with multi-scale reasoning.}
\label{fig_intro}  
\vspace{-0.15in}
\end{figure} 

\section{Introduction}
\label{sec:intro}

Zero-Shot Composed Image Retrieval (ZS-CIR)~\cite{baldrati2023zero, saito2023pic2word,  tang2024context} aims to retrieve the target image from gallery images by integrating information from a reference image and a modification text, without training with annotated triplets data.
In contrast to traditional image retrieval tasks~\cite{i2i, i2i2, t2i, t2i2}, which typically involve a single modality, CIR queries necessitate precise ``editing" to the reference image based on the modification text. Therefore, successfully completing the CIR task entails (1) \textbf{Advanced multimodal composed understanding abilities} to accurately interpret the visual context and user's modification intent in modification text, and (2) \textbf{Robust multimodal reasoning abilities} to implement the modifications appropriately.

As shown in Figure~\ref{fig_intro} (a), previous methods~\cite{saito2023pic2word, baldrati2023zero} primarily propose a textual inversion module to generate pseudo-tokens from the reference image and concatenate it with the modification text. However, these methods still require extensive data for training, and the relatively short length of pseudo-tokens limits the model’s ability to fully understand the reference image. Notably, these statistical methods lack enough logical reasoning for the CIR task.
Recent works~\cite{CIReVL, ldre} leverage Large Language Models (LLMs) to identify the user's modification intent. As shown in Figure~\ref{fig_intro} (b), these methods use pre-trained caption models to generate a caption for the reference image, and then employ the LLM to edit this caption based on the modification text. 
However, the cascading combination of different models introduces several challenges. (1) \textit{Component Incompatibility:} There are domain gaps in language style and way of thought between caption models and LLMs; (2) \textit{Visual Information Loss:} During the caption generation process, some detailed information about the reference image is inevitably lost; (3) \textit{Single-scale Reasoning:} Existing methods focus solely on image-scale reasoning, neglecting fine-grained details; (4) \textit{Insufficient Reasoning:} As a key component, current approaches have not fully leveraged the reasoning capability of LLM.
Therefore, as shown in Figure~\ref{fig_intro}, the aforementioned methods make it hard to preserve the ``golden retriever" characteristics in the reference image.

In this work, we propose \textbf{CoTMR}, a training-free and highly interpretable framework crafted for ZS-CIR with novel Chain-of-Thought (CoT) and Multi-scale Reasoning. As shown in Figure~\ref{fig_intro} (c), instead of relying on the combination of caption models and LLMs, our CoTMR employs the Large Vision-Language Model (LVLM) to achieve unified understanding and reasoning for composed queries. This framework offers several appealing benefits, including rich visual information, unified reasoning, and simplified workflow. Furthermore, we propose a novel CoT method, named \textbf{CIRCoT}, to further enhance the reasoning capability and interpretability of the LVLM in the CIR task. Unlike previous works~\cite{zheng2023ddcot} that entirely delegate the task decomposition process to the model, CIRCoT pre-divides the CIR task into multiple subtasks and allows the model to reason each pre-defined subtask step-by-step. Additionally, a few examples can also be included for reference in CoT~\cite{fewshotcot1}. 
This structured reasoning process not only guides the LVLM through a step-by-step inference process but also provides high-level interpretability, allowing users to intervene for more precise retrieval when necessary.

With this structured reasoning process, we further propose \textbf{Multi-Scale Reasoning} to obtain both the global description and fine-grained details of the target image from the composed query. As shown in Figure~\ref{fig_intro} (c), in addition to reasoning the ``target image caption" at the image scale, we further conduct object-scale reasoning to emphasize key objects and attributes. Notably, aligning with the requirement of CIR, we should not only infer the objects that should be present in the target image (``existent objects") but also naturally take those that should not be present (``nonexistent objects") into account. The existent objects further supplement the target image caption, while nonexistent objects are used to reduce distracting information. Subsequently, \textbf{Multi-Grained Scoring (MGS)} mechanism is designed to enable a precise retrieval process. This module comprehensively considers the characteristics of these multi-grained outputs and separately calculates their similarity scores with the candidate images via CLIP~\cite{clip}. Ultimately, MGS integrates these scores together to achieve a balanced evaluation by rewarding the presence of relevant content while penalizing irrelevant or conflicting content.

Our main contributions can be summarized as follows:
(1) We propose CoTMR, a novel training-free LVLM-based framework for ZS-CIR.
(2) We present multi-scale reasoning and a novel scoring module to provide multi-grained descriptions and evaluations.
(3) We design a novel CIRCoT, which standardizes the LVLM's reasoning process, allowing it to focus on specific goals at each subtask.
(4) Extensive experiments demonstrate that our CoTMR not only significantly outperforms state-of-the-art methods across three prominent benchmarks but also offers appealing interpretability for CIR.

\section{Related Work}

\subsection{Zero-Shot Composed Image Retrieval}

CIR~\cite{TIGR, CoSMo, VAL, clip4cir} integrates concepts from compositional learning~\cite{misra2017red, karthik2022kg} and cross-modal retrieval~\cite{qian2021adaptive,qian2022integrating}. To mitigate the high cost and time-consuming nature of training dataset annotation for CIR, ZS-CIR has recently been introduced. Currently, two prominent directions exist: one approach~\cite{saito2023pic2word, baldrati2023zero, lincir} trains a textual inversion module using only image-caption data, representing the reference image with a single pseudo-token that is then concatenated with the reference caption. This method not only requires training but also is limited by the length of the pseudo-token, which constrains the representation of the reference image. The other approach~\cite{CIReVL, ldre, GRB} forms a training-free method by cascading multiple off-the-shelf tools. It first converts reference image into a textual description using a captioning model and then edits this caption according to the modification text by a LLM. Finally, the edited caption is used to compute CLIP scores with candidate images for retrieval. However, such methods face several challenges, including component incompatibility, visual information loss, and insufficient reasoning. In this work, we propose a unified, training-free, and interpretable framework with CIRCoT and Multi-Scale Reasoning.

\begin{figure*}[t!]
\centering  
\includegraphics[width=0.98\linewidth]{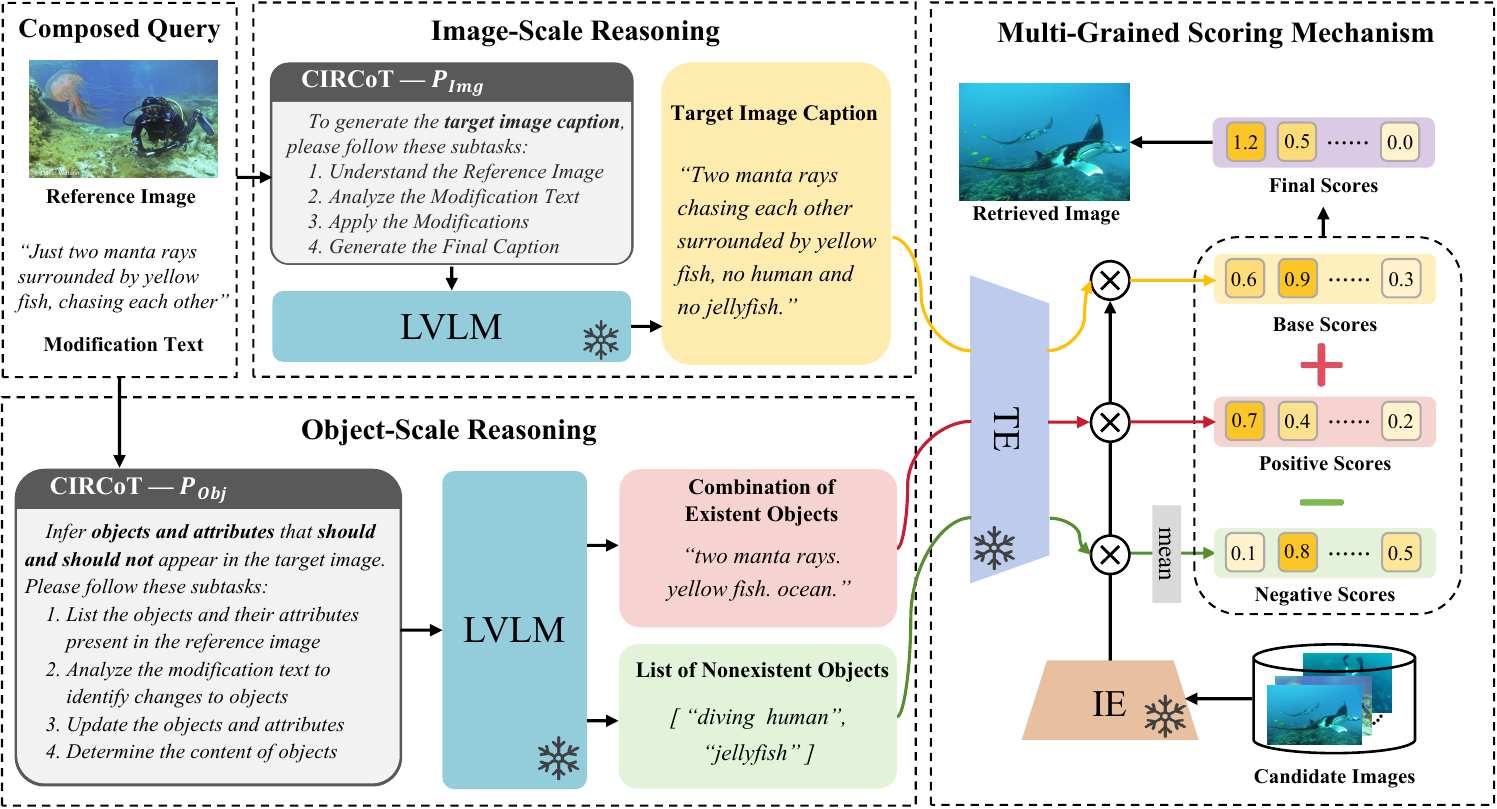} 
\caption{\textbf{Overview architecture of CoTMR:} (1) The LVLM equipped with CIRCoT, $P_{Img}$ and $P_{Obj}$, performs reasoning on the composed query at both image and object scales, to provide multi-grained outputs. (2) The Multi-Grained Scoring Mechanism combines the similarities of the three outputs with candidate images in the CLIP space through a reward-penalty calculation. IE and TE represent the image encoder and text encoder of CLIP, respectively.}
\label{model_art}  
\vspace{-0.1in}
\end{figure*} 

\subsection{Vison-Language Model}

There are two main types of Vision-Language Models (VLMs). The first type, including models like CLIP~\cite{clip} and BLIP~\cite{li2022blip}, is pre-trained on large-scale image-caption datasets, enabling them to map images and text into a shared embedding space for cross-modal retrieval~\cite{bogolin2022cross, roth2022integrating} or open-vocabulary classification~\cite{pratt2023does, udandarao2023sus}. In this work, we use CLIP for the multimodal retrieval process. The second type is LVLM~\cite{dai2024instructblip, liu2023llava15, wang2024qwen2vl}, which are pre-trained to integrate visual information into LLM and are post-trained to align with users. Thus, LVLMs could understand user intent and process various visual tasks, such as image captioning~\cite{agrawal2019nocaps,lin2014coco}, VQA~\cite{hudson2019gqa, goyal2017vqa}, and OCR~\cite{mishra2019ocr, sidorov2020textcaps}. In this work, we utilize the LVLM to inference both the global description and fine-grained details based on composed queries.

\subsection{Chain of Thought}

Recently, zero-shot~\cite{zeroshotcot} and few-shot~\cite{fewshotcot1, fewshotcot2} multi-step reasoning prompts have shown significant enhancement to the reasoning capabilities of LLMs. Consequently, CoT strategy raises increasing research attention and is also extended into multimodal domains. MM-CoT~\cite{zhang2023multimodal} designs a two-stage framework where the model initially learns to generate rationales based on real annotations and then uses all available information to produce the final answer. DDCoT~\cite{zheng2023ddcot} focuses on text understanding, breaking down questions into sub-questions for step-by-step responses. CCoT~\cite{ccdot}, on the other hand, is based on image understanding, generating scene graphs of images to provide answers. However, several works~\cite{cot_survey,sprague2024cot} suggest that CoT seems to work effectively only in some specific domains. In this work, we propose CIRCoT, which pre-divides the task into multiple subtasks and allows the model to reason these subtasks step-by-step.

\section{Methodology}

\subsection{Preliminary}
\label{ZS-CIR_Preliminary}
Given a composed query $Q=\{I_r, T_m\}$, where $I_r$ denotes the reference image and $T_m$ denotes the modification text, and a candidate set $D=\{I_t^1, I_t^2, ..., I_t^{N_D}\}$ consisting of $N_D$ images, the goal of CIR is to identify the $k$ target images from the candidate set $D$ that are most relevant to the query $Q$, with $k \ll N_D$. ZS-CIR further requires that no training data triplets be used. 

Different from the traditional multi-modal retrieval task~\cite{i2i, i2i2, t2i, t2i2}, CIR requires the model to retrieve images that both preserve the key features of the reference image and satisfy the modifications described in the modification text. Successfully completing the CIR task requires: (1) correctly understanding the content of the reference image and the modification text, (2) accurately applying the modifications, and (3) an effective score mechanism for the retrieval. Therefore, CIR methods should possess advanced multimodal composed understanding and reasoning capabilities, as well as a comprehensive score mechanism.

\subsection{Overall Architecture}
Our proposed CoTMR is an effective, training-free and interpretable CIR framework based on public pre-trained VLMs. As shown in Figure~\ref{model_art}, our CoTMR consists of two steps: reasoning the composed query by LVLM and retrieving the target image by CLIP. In the reasoning process, to enhance the interpretability and reliability of reasoning, we first propose CIRCoT, a novel CoT strategy with predefined subtask divisions tailored for CIR. Moreover, we conduct image-scale and object-scale reasoning, both with CIRCoT, to obtain the global description and fine-grained details for the target image. For the retrieval process, we design a novel Multi-Grained Scoring (MGS) mechanism that comprehensively considers the characteristics of the above reasoning outputs at different scales via a reward-penalized formulation. We describe the three modules below.

\subsection{CIRCoT}
\label{Predefined_Sub-Problem_Division}
CIR requires precise understanding and reasoning of the multi-modal composed query, making it a complex task. To achieve a more accurate and reliable reasoning process, we propose using CoT to facilitate multi-step reasoning of LVLM. However, we find that traditional CoT approaches, such as DDCoT~\cite{zheng2023ddcot}, which typically rely on LVLM itself to independently develop problem-solving and task decomposition strategies, tend to work effectively only in a few specific domains~\cite{sprague2024cot}.
Given the certainty of CIR inputs (reference image and modification text) and the clarity of CIR task (editing the reference image according to the modification text), we thus propose CIRCoT, which decomposes the CIR task into multiple subtasks in advance.

\begin{figure}[t!]
\centering  
\includegraphics[width=0.99\linewidth]{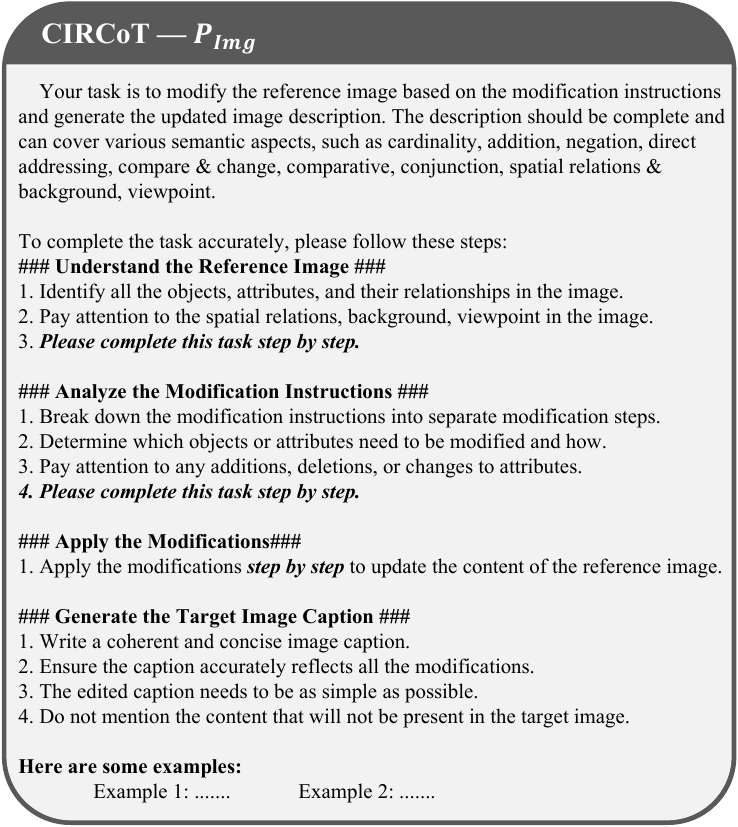} 
\vspace{-0.05in}
\caption{\textbf{Illustration of CIRCoT in image-scale reasoning ($P_{Img}$)}, which includes four predefined subtasks and allows LVLM to reason step-by-step within each subtask. CIRCoT in object-scale reasoning ($P_{Obj}$) follows a similar process (see appendix~\ref{sec_appendix_P_O} for details). 
}
\label{cot_example}  
\vspace{-0.1in}
\end{figure}

As illustrated in Figure~\ref{cot_example}, we divide the task of generating the target image caption using the LVLM into four key subtasks: (1) \textit{Image understanding}; (2) \textit{Modification text understanding}; (3) \textit{Modification implementation} and (4) \textit{Target image caption generation}. These four fundamental subtasks structure the overall reasoning process of the LVLM. For each subtask, we adhere to the traditional CoT approach, enabling the model to reason in a step-by-step manner (as represented by italicized prompts in Figure~\ref{cot_example}). Additionally, we incorporate several step-wise reasoning examples to further stimulate the LVLM's reasoning capability like~\cite{fewshotcot1}.
We emphasize that CIRCoT not only capitalizes on the LVLM's reasoning capability but also offers significant interpretability.
Users can clearly follow the LVLM’s inference process and, if needed, intervene to modify it. More details can be found in Appendix~\ref{sec_appendix_user_interv}.

Combined with the multi-scale reasoning strategy to be introduced, we have two CIRCoT prompts in this work, denoted as $P_{Img}$ and $P_{Obj}$, which are applied at image scale and object scale, respectively. Here, we take $P_{Img}$ as an example shown in Figure~\ref{cot_example}, and $P_{Obj}$ follows a similar process (see Appendix~\ref{sec_appendix_P_O} for details).

\subsection{Mutil-Scale Reasoning}
\label{Mutil-Scale_Reasoning}
When applying LVLM to address ZS-CIR task, a straightforward approach is to directly infer the target image caption based on the composed query. However, this global description presents several challenges: (1) The dense semantic content in the generated caption overshadows the key objects and attributes that need more attention. (2) In complex scenarios, the model may be confused by irrelevant details in the reference image (e.g. the ``human" and ``jellyfish" in Figure~\ref{model_art}). 

To alleviate the negative impact of unclear key features and irrelevant information contained in the global caption, in addition to global caption generation, we propose to reason at the object scale to obtain supplementary fine-grained details. As shown in Figure~\ref{model_art}, at image-scale reasoning, we utilize the LVLM to reason the editing process and generate the target image caption, which is formulated as:
\begin{equation}
    T_{tc} = LVLM(I_r, T_m, P_{Img})
\end{equation}
where $T_{tc}$ is the target image caption, $LVLM(\cdot)$ denotes the reasoning process with LVLM and $P_{Img}$ denotes the CIRCoT prompt at image-scale reasoning.

In object-scale reasoning, we let LVLM focus on specific objects and their attributes, specifying the set of objects that should be present in the target image (\textit{``existent objects''}), and those should not be present (\textit{``nonexistent objects''}). This process is expressed as:
\begin{equation}
    EO, NEO = LVLM(I_r, T_m, P_{Obj})
\end{equation}
Here, $EO = [T_{eo}^{i}]_{i=0}^{L_e}$ denotes the list of \textit{``existent objects''}, where $T_{eo}^{i}$ represents the $i-th$ object that should be present and $L_e$ is the total number of these existent objects. Similarly, $NEO = [T_{neo}^{i}]_{i=0}^{L_u}$ denotes the list of \textit{``nonexistent objects"}. $T_{neo}^{i}$ and $L_u$ represents the $i-th$ nonexistent object and the total number of these objects. $P_{Obj}$ denotes the CIRCoT prompt used at object-scale reasoning.

As shown in Figure~\ref{model_art}, the \textit{``existent objects''} further emphasize the key elements that require extra attention (``Two manta rays", ``yellow fish" and ``ocean"), while the \textit{``nonexistent objects''} mitigate the influence of irrelevant information from the reference image (``human" and ``jellyfish").

\subsection{Multi-Grained Scoring}
\label{Addition-Subtraction_Scoring_Mechanism}
After obtaining the above outputs at multiple scales (target image caption, existent objects, and nonexistent objects), we further design this MGS mechanism to comprehensively consider their impact on the final retrieval process. 

Specifically, as illustrated in Figure~\ref{model_art}, we first compute similarities between the ``target image caption" and candidate images using CLIP as the base scores $S_{base}$:
\begin{equation}
S_{base} = CLIP(T_{tc}, D)
\end{equation}
where $D$ denotes the set of candidate images, and $CLIP(\cdot, \cdot)$ computes the similarity between text and images in the CLIP space. 

At object scale, considering that ``existent objects" typically have inherent correlations, they should be treated as a whole to collectively influence the matching result. Therefore, we concatenate these objects into one string and then compute its similarities with candidate images to obtain positive scores $S_{pos}$:
\begin{equation}
S_{pos} = CLIP(Concat([ T_{eo}^{i} ]_{i=0}^{L_e}), D)
\end{equation}
where $Concat(\cdot)$ denotes the concatenation of strings.

In contrast, ``nonexistent objects" usually have no inherent correlation with each other. Thus, we first calculate their similarities with candidate images individually, and then average their similarities to derive the negative scores $S_{neg}$:
\begin{equation}
S_{neg} = Avg([CLIP(T_{neo}^{i}, D)]_{i=0}^{L_u})
\end{equation}
where $Avg(\cdot)$ denotes the average of scores. This strategy ensures an equal contribution of each undesired object.

Finally, we combine base scores, positive scores, and negative scores using weighted aggregation to obtain the final scores $S$ served as selection criteria:
\begin{equation}
S = S_{base} + \lambda \cdot S_{pos} - \mu \cdot S_{neg}
\end{equation}
where $\lambda$ and $\mu$ are the weights assigned to the positive score, and negative score, respectively. 

Notably, unlike previous works that mostly focused on the updated content of the target image, our MGS mechanism assigns negative scores to objects that should not be present, which is advantageous in better filtering out misleading candidate images. Furthermore, by rewarding the presence of relevant content while penalizing irrelevant content, our score mechanism ensures a much more comprehensive evaluation for CIR.


\begin{table*}[t!]
\centering
\scalebox{0.85}{
\tabcolsep8pt
\begin{tabular}{llcccccccccc}
\toprule
\multirow{2}{*}{Backbone} & \multirow{2}{*}{Method} & \multirow{2}{*}{Training-free} & \multicolumn{2}{c}{Shirt} & \multicolumn{2}{c}{Dress} & \multicolumn{2}{c}{Tops\&Tee} & \multicolumn{3}{c}{Avg.} \\ \cline{4-12} 
 & &  & R@10 & R@50 & R@10 & R@50 & R@10 & R@50 & R@10 & R@50 & R$_{mean}$ \\ \hline
\multirow{6}{*}{ViT-B/32}
 & PALAVRA & \ding{55} & 21.49 & 37.05 & 17.25 & 35.94 & 20.55 & 38.76 & 19.76 & 37.25 & 28.50 \\
 & SEARLE & \ding{55} & 24.44 & 41.61 & 18.54 & 39.51 & 25.70 & 46.46 & 22.89 & 42.53 & 32.71 \\
 & CIReVL & \usym{1F5F8} & {\ul 28.36} & {\ul 47.84} & {\ul 25.29} & {\ul 46.36} & {\ul 31.21} &  {\ul 53.85} & {\ul 28.29} & {\ul 49.35} & {\ul 38.82} \\
 & LDRE & \usym{1F5F8} & 27.38 & 46.27 & 19.97 & 41.84 & 27.07 & 48.78 & 24.81 & 45.63 & 35.22 \\
 & \textbf{CoTMR} & \usym{1F5F8} & \textbf{33.42} & \textbf{53.93} & \textbf{31.09} & \textbf{54.54} & \textbf{38.40} & \textbf{61.14} & \textbf{34.30} & \textbf{56.54} & \textbf{45.42} \\ \hline
\multirow{4}{*}{ViT-L/14} & Pic2Word & \ding{55} & 26.2 & 43.6 & 20.00 & 40.2 & 27.9 & 47.40 & 24.70 & 43.70 & 34.20 \\
 & SEARLE & \ding{55} & 26.89 & 45.58 & 20.48 & 43.13 & 29.32 & 49.97 & 25.56 & 46.23 & 35.89 \\
  & CIReVL & \usym{1F5F8} & 29.49 & 47.40 & 24.79 & 44.76 & 31.36 &  53.65 & 28.55 & 48.57 & 38.56 \\
 & LDRE & \usym{1F5F8} & {\ul 31.04} & {\ul 51.22} & {\ul 22.93} & {\ul 46.76} & {\ul 31.57} & {\ul 53.64} & {\ul 28.51} & {\ul 50.54} & {\ul 39.52} \\
 & \textbf{CoTMR} & \usym{1F5F8} & \textbf{35.43} & \textbf{54.91} & \textbf{31.18} & \textbf{55.04} & \textbf{38.55} & \textbf{61.33} & \textbf{35.05} & \textbf{57.09} & \textbf{46.50} \\ \hline
\multirow{2}{*}{ViT-G/14} & CIReVL & \usym{1F5F8} & 33.71 & 51.42 & 27.07 & 49.53 & 35.80 & 56.14 & 32.19 & 52.36 & 42.27 \\
& LDRE & \usym{1F5F8} & {\ul 35.94} & {\ul 58.58} & {\ul 26.11} & {\ul 51.12} & {\ul 35.42} & {\ul 56.67} & {\ul 32.49} & {\ul 55.46} & {\ul 43.97} \\
 & \textbf{CoTMR} & \usym{1F5F8} & \textbf{38.32} & \textbf{62.24} & \textbf{34.51} & \textbf{57.36} & \textbf{41.90} & \textbf{64.30} & \textbf{38.25} & \textbf{61.32} & \textbf{49.78} \\ \bottomrule
\end{tabular}}
\vspace{-0.08in}
\caption{\textbf{Comparison with the state-of-the-art methods on the Fashion-IQ dataset.} R$_{mean}$ indicates the average results across all the metrics. The best results are in boldface, while the second-best results are underlined.}
\label{tab_fashioniq_main}
\vspace{-0.08in}
\end{table*}

\begin{table*}[]
\centering
\scalebox{0.85}{
\tabcolsep5.2pt
\begin{tabular}{clc|cccc||ccccccc|c}
\toprule
\multicolumn{3}{c|}{Benchmark} & \multicolumn{4}{c||}{\textbf{CIRCO}} & \multicolumn{8}{c}{\textbf{CIRR}} \\ \midrule
\multicolumn{3}{c|}{Metric} & \multicolumn{4}{c||}{mAP@k} & \multicolumn{4}{c|}{Recall@k} & \multicolumn{3}{c|}{Recall$_{sub}$@k} & \multirow{2}{*}{Avg.} \\
\multicolumn{1}{l}{Backbone} & Method & Training-free & k=5 & k=10 & k=25 & k=50 & k=1 & k=5 & k=10 & \multicolumn{1}{c|}{k=50} & k=1 & k=2 & k=3 &  \\ \midrule
\multirow{7}{*}{ViT-B/32} 
 & PALAVRA & \ding{55} & 4.61 & 5.32 & 6.33 & 6.80 & 16.62 & 43.49 & 58.51 & \multicolumn{1}{c|}{83.95} & 41.61 & 65.30 & 80.94 & 42.55 \\
 & SEARLE & \ding{55} & 9.35 & 9.94 & 11.13 & 11.84 & 24.00 & 53.42 & 66.82 & \multicolumn{1}{c|}{89.78} & 54.89 & 76.60 & 88.19 & 54.15 \\
 & CIReVL & \usym{1F5F8} & 14.94 & 15.42 & 17.00 & 17.82 & 23.94 & 52.51 & 66.00 & \multicolumn{1}{c|}{86.95} & 60.17 & 80.05 & 90.19 & 56.34 \\
 & LDRE & \usym{1F5F8} & {\ul 17.96} & {\ul 18.32} & {\ul 20.21} & {\ul 21.11} & {\ul 25.69} & {\ul 55.13} & {\ul 69.04} & \multicolumn{1}{c|}{{\ul 89.90}} & {\ul 60.53} & {\ul 80.65} & {\ul 90.70} & {\ul 57.83} \\
 & \textbf{CoTMR} & \usym{1F5F8} & \textbf{22.23} & \textbf{22.78} & \textbf{24.68} & \textbf{25.74} & \textbf{31.50} & \textbf{60.80} & \textbf{73.04} & \multicolumn{1}{c|}{\textbf{91.06}} & \textbf{66.61} & \textbf{84.50} & \textbf{92.55} & \textbf{63.71} \\ \midrule
\multirow{7}{*}{ViT-L/14} & Captioning & \ding{55} & 1.65 & 1.96 & 2.42 & 2.71 & 4.05 & 15.88 & 25.69 & \multicolumn{1}{c|}{49.21} & 20.87 & 40.60 & 60.89 & 18.37 \\
 & Pic2Word & \ding{55} & 8.72 & 9.51 & 10.64 & 11.29 & 23.90 & 51.70 & 65.30 & \multicolumn{1}{c|}{87.80} & - & - & - & - \\
 & SEARLE & \ding{55} & 11.68 & 12.73 & 14.33 & 15.12 & 24.24 & 52.48 & 66.29 & \multicolumn{1}{c|}{{\ul 88.84}} & 53.76 & 75.01 & 88.19 & 53.12 \\
 & CIReVL & \usym{1F5F8} & 18.57 & 19.01 & 20.89 & 21.80 & 24.55 & 52.31 & 64.92 & \multicolumn{1}{c|}{86.34} & 59.54 & 79.88 & 89.69 & 55.92 \\
 & LDRE & \usym{1F5F8} & {\ul 23.35} & {\ul 24.03} & {\ul 26.44} & {\ul 27.50} & {\ul 26.53} & {\ul 55.57} & {\ul 67.54} & \multicolumn{1}{c|}{88.50} & {\ul 60.43} & {\ul 80.31} & {\ul 89.90} & {\ul 58.00} \\
 & \textbf{CoTMR} & \usym{1F5F8} & \textbf{27.61} & \textbf{28.22} & \textbf{30.61} & \textbf{31.70} & \textbf{35.02} & \textbf{64.75} & \textbf{76.18} & \multicolumn{1}{c|}{\textbf{92.51}} & \textbf{69.39} & \textbf{85.75} & \textbf{93.33} & \textbf{67.07} \\ \midrule
\multirow{3}{*}{ViT-G/14} & CIReVL & \usym{1F5F8} & 26.77 & 27.59 & 29.96 & 31.03 & 34.65 & 64.29 & 75.06 & \multicolumn{1}{c|}{91.66} & 67.95 & 84.87 & 93.21 & 66.12 \\
 & LDRE & \usym{1F5F8} & {\ul 31.12} & {\ul 32.24} & {\ul 34.95} & {\ul 36.03} & {\ul 36.15} & {\ul 66.39} & {\ul 77.25} & \multicolumn{1}{c|}{{\ul 93.95}} & {\ul 68.82} & {\ul 85.66} & {\ul 93.76} & {\ul 67.60} \\
 & \textbf{CoTMR} & \usym{1F5F8} & \textbf{32.23} & \textbf{32.72} & \textbf{35.60} & \textbf{36.83} & \textbf{36.36} & \textbf{67.52} & \textbf{77.82} & \multicolumn{1}{c|}{\textbf{93.99}} & \textbf{71.19} & \textbf{86.34} & \textbf{93.87} & \textbf{69.36} \\ \bottomrule
\end{tabular}}
\vspace{-0.08in}
\caption{\textbf{Comparison  with the state-of-the-art methods on CIRCO and CIRR test sets.} Avg. indicates the average results of Recall@5 and Recall$_{sub}$@1.  The best results are in boldface, while the second-best results are underlined.}
\label{tab_cirr_circo_main}
\vspace{-0.1in}
\end{table*}

\section{Experiments}

\subsection{Implementation Details}
For the LVLM, we use Qwen2-VL-72B~\cite{wang2024qwen2vl}. For the retrieval model, we experiment with different CLIP variants, including ViT-B/32, ViT-L/14, and ViT-G/14 CLIP from OpenCLIP~\cite{Openclip}. The hyperparameter $\lambda$ and $\mu$ are set to 1 and 0.5 for the FashionIQ dataset, 1 and 0.3 for the CIRR dataset, and 0.5 and 0.3 for the CIRCO dataset, respectively. The entire model is implemented using PyTorch~\cite{paszke2019pytorch} on 8 NVIDIA A800 GPUs.

\subsection{Baselines}
We use the ``image-only" and ``text-only" to denote directly performing retrieval with CLIP using only the reference image and modification text. PALAVRA~\cite{PALAVRA}, Pic2Word~\cite{saito2023pic2word}, SEARLE~\cite{baldrati2023zero} are the textual inversion methods either designed or adapted for ZS-CIR. CIReVL~\cite{CIReVL} and LDRE~\cite{ldre} are LLM-based, training-free methods that cascade captioning models and LLMs to generate textual descriptions of the target image. Among them, CIReVL is most similar to our method and serves as the most direct baseline for CoTMR.

\subsection{Datasets and Evaluation Metrics}
We make performance evaluations on three CIR benchmarks, including a fashion-domain dataset \textbf{Fashion-IQ} \cite{HuiWu2019FashionIA}, as well as two open-domain datasets \textbf{CIRR}~\cite{CIRPLANT} and \textbf{CIRCO}~\cite{baldrati2023zero}. 
FashionIQ contains garment images that can be divided into three categories: dress, shirt, and toptee. CIRR is the first natural image dataset designed specifically for CIR. CIRCO is based on real-world images from the COCO 2017 unlabeled set~\cite{lin2014microsoft} and is the first dataset for CIR to provide multiple ground truths.

For FashionIQ, we adopt Recall@K (R@K) as the evaluation metric, which refers to the fraction of queries for which the correct item is retrieved among the top K results. We also report $R_{mean}$, the mean of all R@K values, to evaluate the overall retrieval performance. For CIRR, beside Recall@K, we additionally report Recall$_{subset}$@K and the average score of Recall@5 and Recall$_{subset}$@1 as in \cite{CIRPLANT}. For CIRCO, since there are multiple positives, we use the mean average precision@k (mAP@k) as the metric.


\subsection{Comparison with Bselines}
\textbf{Fashion-IQ.} 
Table~\ref{tab_fashioniq_main} presents the comparative results on the  Fashion-IQ dataset. Based on the results, we have the following observations: (1) Compared to pseudo-word-based methods such as SEARLE, CoTMR achieves impressive performance across multiple metrics even without any training. This indicates that generating captions for the target image using LVLM provides semantic information that is more suitable for CLIP’s text encoding than concatenating modification texts with pseudo-words. (2) Compared to LDRE, which also uses large language models to model the modified images, our approach achieves significant improvements. This is attributed to CoTMR’s superior preservation of image semantics, more refined reasoning process, and finer-grained feature recognition. (3) Across all metrics with different CLIP backbones, CoTMR consistently outperforms all baseline methods. Using ViT-B/32 as an example, our method relatively outperforms LDRE by 9.49\% in average R@10 and 10.91\% in average R@50. These results strongly support CoTMR’s effectiveness.

\noindent \textbf{CIRR.} When applied to the open-domain dataset CIRR,  CoTMR still shows compelling results, summarized in the right section of Table~\ref{tab_cirr_circo_main}. Based on the results, we have the following observations: (1) Notably, the CIRR dataset is quite noisy, with minimal correlation between the reference image and the target image, especially compared to the modification text. Therefore, CoTMR's ability to capture rich information from reference images also means it may receive more distracting information. Despite this challenge, CoTMR consistently outperforms all baseline metrics across all CLIP architectures. These findings highlight the robustness of our approach, demonstrating its ability to deliver significant results even in the presence of noisy data and its adaptability across diverse scenarios.
(2) CIRR also provides another evaluation, where the task is to retrieve the correct image from six curated samples. In this evaluation, our approach also significantly surpasses previous methods (Our method outperforms LDRE by 6.08\% in Recall$_{sub}$@1 when using ViT-B/32 CLIP). This shows the versatility of our method, enabling it to perform well across different contexts. (3) Using ViT-L/14 as an example, compared to our most direct baseline model, CIReVL, our method achieves significant improvement in Recall@5 and Recall$_{sub}$@1 by 12.44\% and 9.85\% respectively. This further demonstrates the effectiveness of our proposed modules, such as CIRCoT and multi-scale reasoning.

\begin{table}[t!]
\centering
\scalebox{0.9}{
\tabcolsep2pt
\begin{tabular}{lcccccc}
\toprule
\multicolumn{1}{l|}{Benchmark} & \multicolumn{2}{c}{FashionIQ-Avg} & \multicolumn{4}{c}{CIRCO} \\ \midrule
\multicolumn{1}{l|}{Metric} & \multicolumn{2}{c}{Recall@k} & \multicolumn{4}{c}{mAP@k} \\
\multicolumn{1}{l|}{Method} & k=10 & k=50 & k=5 & k=10 & k=25 & k=50 \\ \midrule
\multicolumn{7}{l}{\textbf{A.} Multi-Grained Scoring} \\ \midrule
\multicolumn{1}{l|}{\textbf{A.1} Base} & 33.99 & 56.34 & 26.40 & 27.98 & 30.35 & 31.43 \\
\multicolumn{1}{l|}{\textbf{A.2} Pos + Neg} & 30.50 & 52.65 & 14.92 & 16.49 & 18.33 & 19.12 \\
\multicolumn{1}{l|}{\textbf{A.3} Base + Pos} & 35.62 & 58.39 & 27.54 & 29.59 & 32.24 & 33.26 \\
\multicolumn{1}{l|}{\textbf{A.4} Base + Neg} & 34.42 & 56.95 & 27.28 & 28.30 & 30.63 & 31.82 \\ 
\multicolumn{1}{l|}{\textbf{A.5 Full}} & \textbf{37.72} & \textbf{60.92} & \textbf{28.87} & \textbf{30.61} & \textbf{33.30} & \textbf{34.32}  \\ \hline
\multicolumn{7}{l}{\textbf{B.} Chain of Thought} \\ \midrule
\multicolumn{1}{l|}{\textbf{B.1} No COT} & 31.03 & 51.01 & 20.07 & 21.12 & 23.43 & 24.44 \\
\multicolumn{1}{l|}{\textbf{B.2} DDCOT} & 29.21 & 48.25 & 17.41 & 18.84 & 21.25 & 22.17 \\
\multicolumn{1}{l|}{\textbf{B.3} ZS\_CIRCoT} & 33.41 & 53.50 & 23.54 & 24.88 & 27.42 & 28.40 \\ 
\multicolumn{1}{l|}{\textbf{B.4 CIRCoT}} & \textbf{33.99} & \textbf{56.34} & \textbf{26.40} & \textbf{27.98} & \textbf{30.35} & \textbf{31.43} \\ \midrule
\multicolumn{7}{l}{\textbf{C.} Scoring for Objects} \\ \midrule
\multicolumn{1}{l|}{\textbf{C.1} Pos + mean} & 35.16 & 56.48 & 28.57 & 30.46 & 33.00 & 34.14 \\
\multicolumn{1}{l|}{\textbf{C.2} Neg + concat} & 37.25 & 59.29 & 28.71 & 30.46 & 33.04 & 34.06 \\
\multicolumn{1}{l|}{\textbf{C.3 Normal}} & \textbf{37.72} & \textbf{60.92} & \textbf{28.87} & \textbf{30.61} & \textbf{33.30} & \textbf{34.32} \\ \hline
\multicolumn{7}{l}{\textbf{D.} Scale of LVLM} \\ \midrule
\multicolumn{1}{l|}{\textbf{D.1} Qwen2-VL-2B} & 20.27 & 36.68 & 4.90 & 5.50 & 6.15 & 6.34 \\
\multicolumn{1}{l|}{\textbf{D.2} Qwen2-VL-7B} & 33.35 & 54.02 & 16.10 & 17.05 & 19.45 & 19.41 \\
\multicolumn{1}{l|}{\textbf{D.3 Qwen2-VL-72B}} & \textbf{37.72} & \textbf{60.92} & \textbf{28.87} & \textbf{30.61} & \textbf{33.30} & \textbf{34.32} \\ \bottomrule
\end{tabular}}
\caption{\textbf{Ablation study results for the proposed components on Fashion-IQ val set and CIRCO val sets.} All experiments are performed with the ViT-G/14 CLIP model.}
\label{tab_ablation}
\vspace{-0.1in}
\end{table}

\noindent \textbf{CIRCO.} In the left section of Table~\ref{tab_cirr_circo_main}, we present the competitive results of CoTMR. Based on the results, we make the following observations: (1) Since CIRCO uses mAP as the evaluation metric, the incorrect selection of negative samples has a significant impact on the results. CoTMR, by introducing a negative scoring mechanism, effectively eliminates incorrect samples, achieving optimal performance across multiple metrics. 
(2) Thanks to multi-scale reasoning and the pre-defined subtask decomposition, CoTMR shows substantial improvements over LLM-based methods such as CIReV and LDRE across several metrics. Using ViT-B/32 as an example, our method relatively outperforms LDRE by 4.27\% and CIReVL by 7.27\% in mAP@5. This further demonstrates the effectiveness of CoTMR for ZS-CIR.

\section{Ablation Study}
\textbf{Effects of Multi-Grained Scoring Mechanism.} In Table~\ref{tab_ablation} A, we investigate the impact of multi-scale reasoning and multi-grained scoring mechanism. Our observations are as follows: (1) Compared to using only the base score $S_{base}$ (\textbf{A.1}), only rely on the object-level reasoning output (\textbf{A.2} $S_{pos}$ and $S_{neg}$) leads to a significant decline of CoTMR’s performance. This indicates that the logical relationships between objects captured in the image-level reasoning are still essential for effective retrieval. (2) However, when the base score is combined with either the positive score (\textbf{A.3}) or the negative score (\textbf{A.4}), the model’s performance improves. This further confirms the effectiveness of object-level reasoning. (3) When both $S_{pos}$ and $S_{neg}$ are used with $S_{base}$ (\textbf{A.5}), the performance significantly improves compared to using only one of them. This suggests that $S_{pos}$ and $S_{neg}$ have complementary roles in the retrieval process.

\noindent \textbf{Effects of CIRCoT.} In Table~\ref{tab_ablation} B, taking image-level reasoning as an example, we compare four different approaches for subtask decomposition: without using CoT (\textbf{B.1}), autonomously decompose the sub-problem by LVLM (\textbf{B.2}, DDCoT~\cite{zheng2023ddcot}), CIRCoT without examples (\textbf{B.3}) and our proposed CIRCoT (\textbf{B.3}). The results show that DDCoT caused slight performance degradation compared to not using CoT, indicating that the reasoning process constructed by the model itself may cause confusion in the CIR task. In contrast, when the CIR task is decomposed in advance, the model's performance improves significantly, which clearly shows the effectiveness of the predefined subtasks. Moreover, adding step-wise reasoning examples also provides assistance in helping the model understand task requirements and the reasoning process. For more analysis on efficiency, please refer to Appendix~\ref{sec_appendix_efficiency}.

\begin{figure}[]
\centering  
\includegraphics[width=0.99\linewidth]{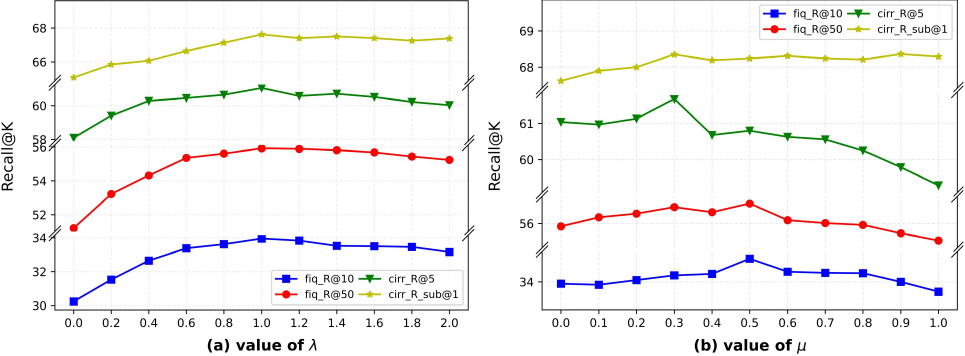} 
\vspace{-0.05in}
\caption{\textbf{Ablation study on the value of $\lambda$ and $\mu$ on Fashion-IQ val set and CIRR val set.} All experiments are performed with the ViT-B/32 CLIP model.}
\label{fig_ablation}  
\vspace{-0.1in}
\end{figure}

\noindent \textbf{Scoring Methods for Objects:} In Table~\ref{tab_ablation} C, we conduct experiment on two different scoring strategies for $S_{pos}$ and $S_{neg}$: (\textbf{C.1}) computing individual scores for each object in 
$EO$ and then averaging them, and (\textbf{C.2}) concatenating all objects in $NEO$ for scoring. Our observations indicate that scoring objects in $EO$ individually results in a significant performance decline compared to the full model. We assume that because of the intrinsic correlations among ``existent objects"—treating them separately overlooks these dependencies. On the other hand, concatenating the uncorrelated objects in $NEO$ may introduce potential bias that causes the decline in performance, as the overall score may be disproportionately affected by certain individual objects.

\noindent \textbf{Scale of LVLM:} As a core component, the performance of the LVLM directly impacts the overall effectiveness. We conduct experiments in Table~\ref{tab_ablation} D, using Qwen2-VL models of different scales: 2-billion parameter (\textbf{D.1}), 7-billion parameter (\textbf{D.2}), and 72-billion parameter (\textbf{D.3}). The results show a sharp decline in performance as the parameter size decreases. However, we observed that with the 7B model, our method achieves relatively satisfactory performance, particularly in tasks like Fashion-IQ, which heavily depend on the reference image. This suggests that the 7B model can adequately comprehend both the visual content of the image and the user’s intent to a certain extent.

\noindent \textbf{Impact of hyperparameter $\lambda$ and $\mu$:} To analyze the sensitivity of the hyperparameters in CoTMR, we conduct controlled experiments as shown in Figure~\ref{fig_ablation}. First, we set $\mu$ to 0 to better demonstrate the effect of $\lambda$. As shown in Figure~\ref{fig_ablation} (a), when the value of $\lambda$ increases from 0, all four metrics show a rapid rise, stabilizing and slightly declining when $\lambda$ reaches 1. Next, we fix $\lambda$ at 1 to explore the effect of $\mu$. As shown in Figure~\ref{fig_ablation} (b), for the Fashion-IQ dataset, the impact of $\mu$ is relatively mild, with the metrics reaching their peak at $\mu = 0.5$. However, for the CIRR dataset, due to the higher noise in the data, increasing $\mu$ too much leads to a significant drop in the R@5 metric. Therefore, for CIRR, the best average performance is achieved when $\mu = 0.3$. 

\begin{figure}[t!]
\centering  
\includegraphics[width=0.99\linewidth]{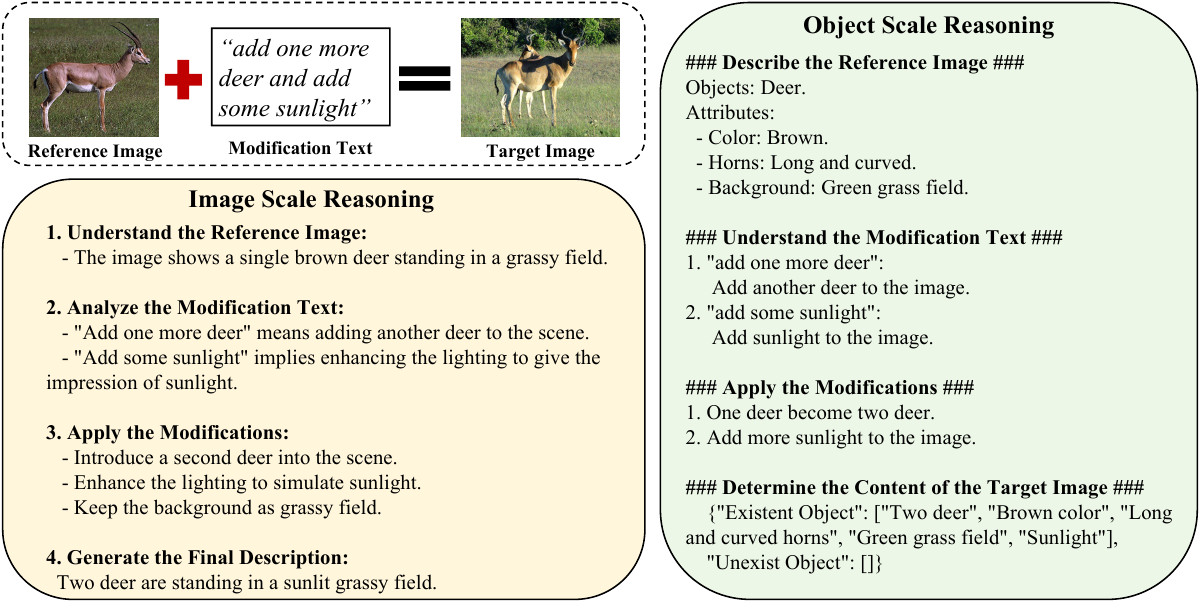} 
\vspace{-0.05in}
\caption{\textbf{An example of a reasoning process with CIRCoT from CIRR val set.} The LVLM focuses on specific objectives in each subtask within CIRCoT and gradually completes the overall reasoning goal.}
\label{qua_cot}  
\vspace{-0.1in}
\end{figure}

\section{Qualitative Results}
\textbf{Reasoning process with CIRCoT:} In Figure~\ref{qua_cot}, we illustrate the reasoning process generated by the LVLM when using CIRCoT at both image and object scale. During image-scale reasoning, the LVLM analyzes the global content of the reference image to ensure comprehensive information coverage. By incrementally breaking down the modification text and executing the modification process, each user modification intent is accurately and completely executed. At object-scale reasoning, the LVLM focuses on the objects and their attributes in the reference image, accurately reasoning which objects and attributes should or should not be present by executing the modification process step-by-step. As a result, the LVLM successfully noticed the key object, i.e., "\textit{long and curved horns}". This predefined structured reasoning process standardizes the model's reasoning path, preventing user modification intents from being overlooked or incorrectly propagated.

\begin{figure}[t!]
\centering  
\includegraphics[width=0.99\linewidth]{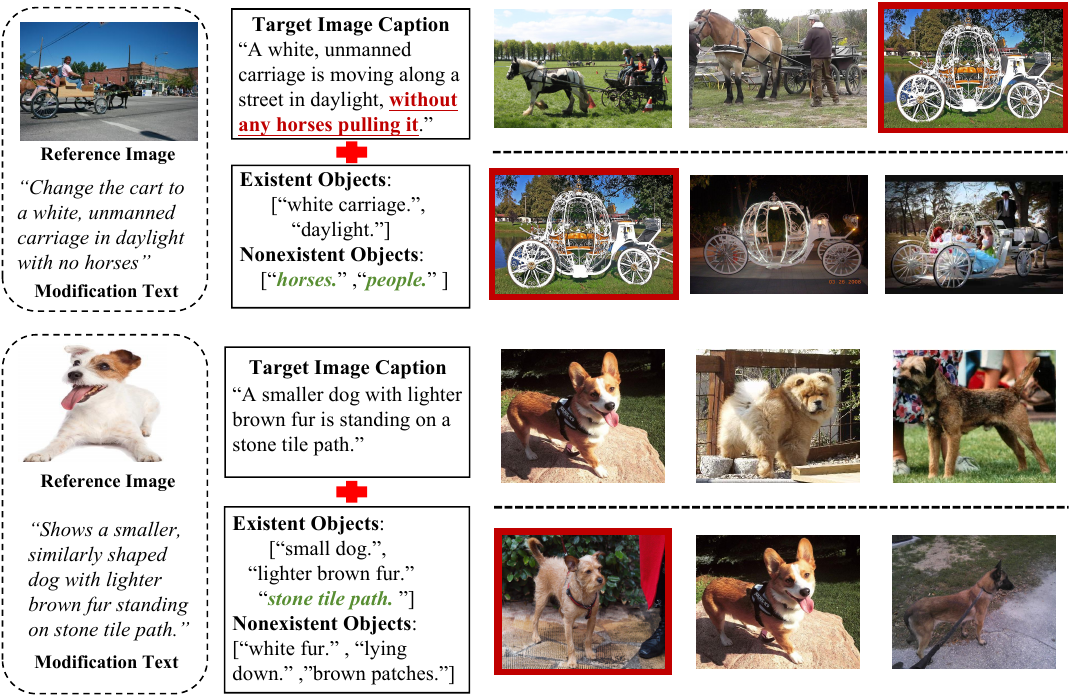} 
\vspace{-0.05in}
\caption{\textbf{Successful retrieval examples with muti-scale reasoning from CIRR val set.} The ground-truth image is highlighted with the red box. Red underlined text indicates distracting information that causes mistake retrieval, while green italicized text represents key objects that help in correct retrieval.}
\label{qua_retrieval}  
\vspace{-0.1in}
\end{figure} 

\noindent \textbf{Examples of successful retrieval.} Figure~\ref{qua_retrieval} visualizes cases where the combination of image-scale and object-scale reasoning leads to successful retrievals. In the first example, the target image caption includes ``\textit{without any horses}", which, while meeting user requirements, is detrimental to CLIP retrieval. However, the undesired objects (``\textit{horses}" and ``\textit{people}") identified through object-scale reasoning eliminates this interference, successfully retrieving the target image. In the second example, we observe that the retrieval results initially overlooked the attribute ``\textit{stone tile path}". Object-scale reasoning, however, highlighted this attribute with existent objects, leading to the successful retrieval of the target image. These examples clearly demonstrate that object-scale reasoning can supplement emphasis and eliminate distracting information. More examples can be found in Appendix~\ref{app_sec:Qualitative}.

\section{Conclusion and Future Work}

In this work, we propose CoTMR, an effective, training-free, and interpretable method for ZS-CIR. It provides a unified understanding and reasoning framework for composed queries, utilizing a step-by-step process guided by CIRCoT. To incorporate fine-grained details, multi-scale reasoning (alongside a novel scoring mechanism) is devised for multi-grained generation and evaluation. Extensive experiments demonstrate the effectiveness of our CoTMR. Moreover, our CoTMR also offers appealing interpretability for user intervention.
However, there are still room for improvement, e.g., designing more suitable CoT modules, or exploiting fine-grained CLIP~\cite{fineclip} or open-vocabulary object detection models~\cite{glip,gdino} for scoring mechanism. We leave these directions for future exploration.

{
    \small
    \bibliographystyle{ieeenat_fullname}
    \bibliography{main}
}
\clearpage
\setcounter{page}{1}
\maketitlesupplementary

\begin{figure}[]
\centering  
\includegraphics[width=1.0\linewidth]{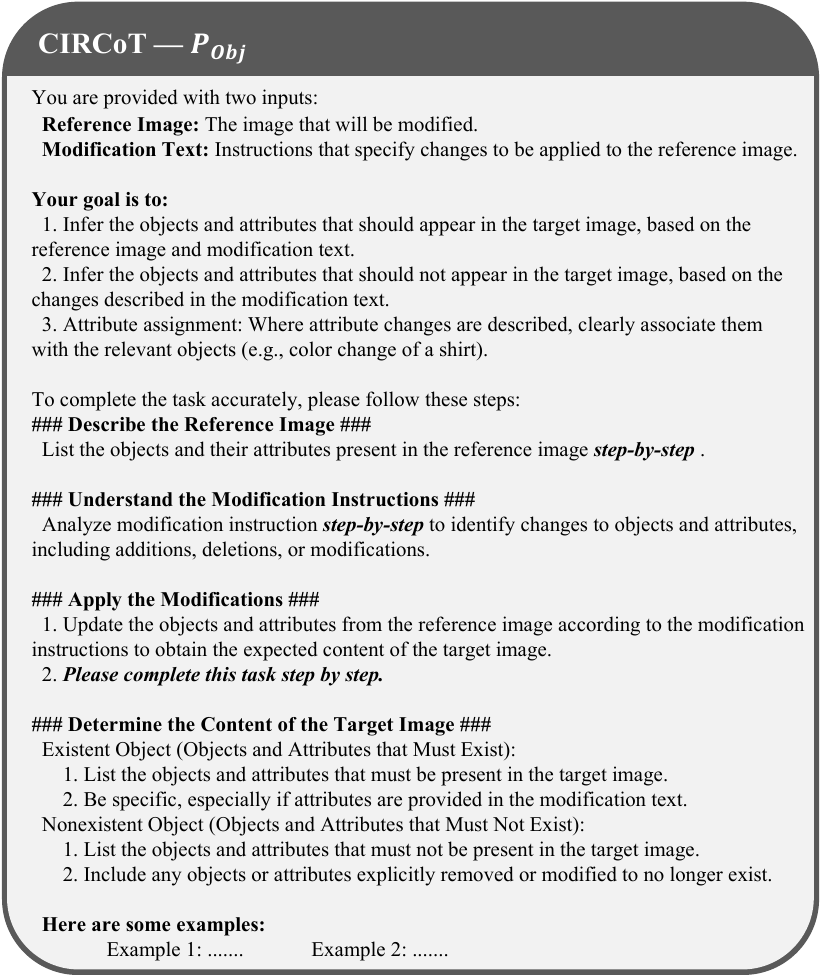} 
\caption{\textbf{Illustration of CIRCoT in object-scale reasoning ($P_O$)}, which includes four predefined subtasks and allows LVLM to reason step-by-step within each subtasks.}
\label{appendix_P_O}  
\vspace{-0.1in}
\end{figure}

\section{CIRCoT in Object-Scale Reasoning}
\label{sec_appendix_P_O}
In Figure~\ref{appendix_P_O}, we show the details of CIRCoT used in object-scale reasoning ($P_O$). Similar to image-scale reasoning, we divide the task of generating ``existent objects" and ``nonexistent objects" into four subtasks: (1) \textit{Describe the Reference Image}; (2) \textit{Understand the Modification Instructions}; (3) \textit{Apply the Modifications} and (4) \textit{Determine the Content of the Target Image}. These four fundamental tasks serve to structure the overall reasoning process of the LVLM. For each subtask, we allow the model to reason step-by-step. Additionally, we add several reasoning examples to further stimulate the model's reasoning capabilities.

\section{Efficiency Analysis}
\label{sec_appendix_efficiency}

We conducted a comprehensive timing analysis to measure the average computational overhead for processing one composed query at the image scale under three different configurations. The results show that the baseline without CoT requires \textbf{0.86s}, while implementations with DDCoT~\cite{zheng2023ddcot} and CIRCoT take \textbf{3.145s} and \textbf{3.183s}, respectively. We have the following observations: \textbf{(1)} While CIRCoT introduces additional computational overhead compared to the non-CoT baseline, this trade-off is justified by the substantial performance improvements it delivers. \textbf{(2)} Notably, CIRCoT achieves significantly enhanced model performance while maintaining comparable computational efficiency to DDCoT, with only a marginal increase in processing time. Future research directions could focus on optimizing CIRCoT's computational efficiency while preserving its superior performance characteristics.

\begin{figure}[t!]
\centering  
\includegraphics[width=1.0\linewidth]{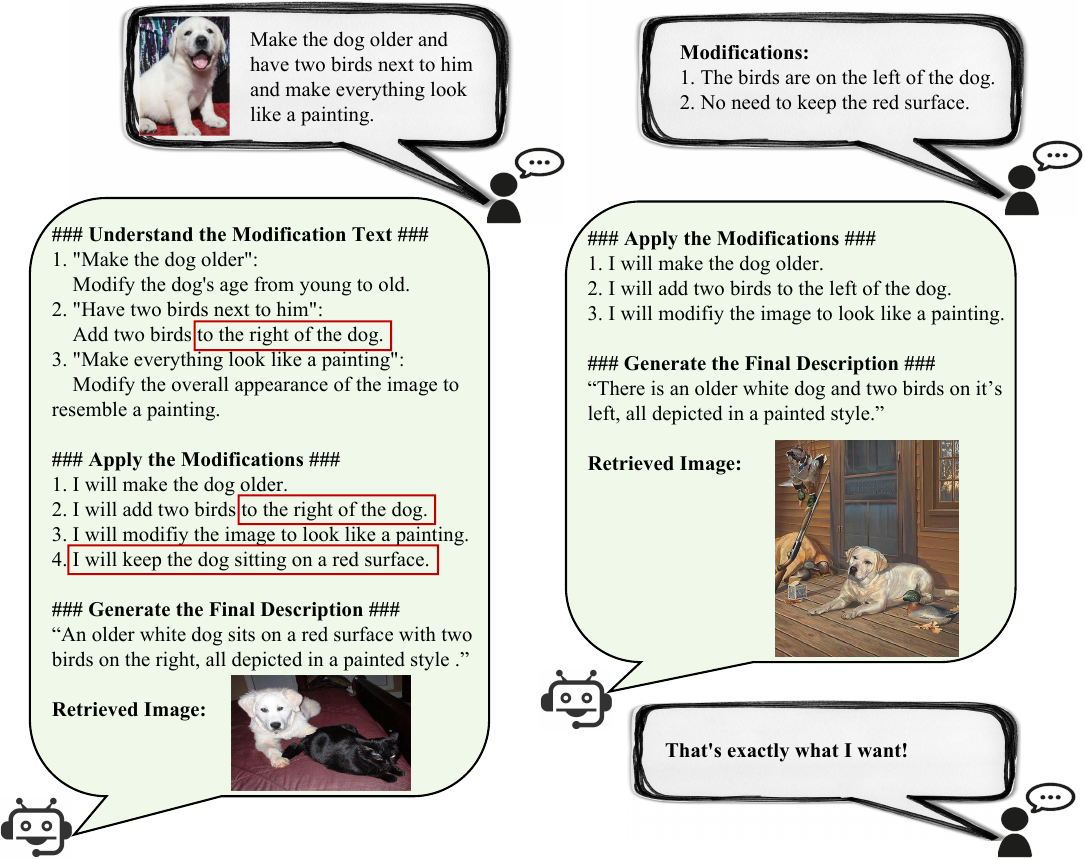} 
\caption{\textbf{The demonstration of making user interventions to enhance ZS-CIR performance with CoTMR.} For instance, by fixing the mistakes in the reasoning process, users are able to correctly retrieve the desired image with further dialogue.}
\label{appendix_user_interv}  
\vspace{-0.15in}
\end{figure}

\begin{table*}[]
\centering
\scalebox{0.85}{
\tabcolsep8pt
\begin{tabular}{ll|ccccccccc}
\toprule
\multirow{2}{*}{} & \multirow{2}{*}{Method} & \multicolumn{2}{c}{Shirt} & \multicolumn{2}{c}{Dress} & \multicolumn{2}{c}{Tops\&Tee} & \multicolumn{3}{c}{Avg.} \\ \cline{3-11} 
 &  & R@10 & R@50 & R@10 & R@50 & R@10 & R@50 & R@10 & R@50 & R$_{mean}$ \\ \midrule
\multirow{2}{*}{One process} & image-scale & 29.44 & 47.11 & 26.82 & 48.79 & 32.33 & 54.77 & 29.53 & 50.22 & 39.87 \\
 & multi-scale & 32.14 & 50.25 & 30.04 & 51.81 & 35.34 & 59.10 & 32.50 & 53.72 & 43.11 \\ \midrule
\multirow{2}{*}{Two processes} & image-scale & 30.03 & 48.58 & 26.57 & 48.69 & 34.12 & 56.35 & 30.24 & 51.21 & 40.72 \\
 & multi-scale & \textbf{33.42} & \textbf{53.93} & \textbf{31.09} & \textbf{54.54} & \textbf{38.40} & \textbf{61.14} & \textbf{34.30} & \textbf{56.54} & \textbf{45.42} \\ \bottomrule
\end{tabular}}
\caption{\textbf{Ablation study on the impact of process quantity in Multi-Scale Reasoning on Fashion-IQ dataset.} All experiments are performed with the ViT-B/32 CLIP model.}
\label{appdix_process}
\vspace{-0.08in}
\end{table*}

\begin{table*}[]
\centering
\scalebox{0.85}{
\tabcolsep8pt
\begin{tabular}{ll|cccc|ccc|c}
\toprule
 &  & \multicolumn{4}{c|}{ { Recall@k}} & \multicolumn{3}{c|}{{ Recall$_{sub}$@k}} &  \\  \cline{3-9}
\multirow{-2}{*}{} & \multirow{-2}{*}{Method} & k=1 & k=5 & k=10 & k=50 & k=1 & k=2 & k=3 & \multirow{-2}{*}{Avg.} \\ \midrule
 & image-scale & 30.76 & 59.01 & 70.75 & 90.34 & 66.08 & 83.74 & 91.96 & 62.54 \\
\multirow{-2}{*}{One process} & multi-scale & 29.56 & 58.69 & 70.27 & 89.72 & 65.61 & 83.52 & 91.68 & 62.15 \\ \midrule
 & image-scale & 30.11 & 58.10 & 70.58 & 89.95 & 65.08 & 83.07 & 91.41 & 61.59 \\
\multirow{-2}{*}{Two processes} & multi-scale & \textbf{31.88} & \textbf{61.27} & \textbf{72.90} & \textbf{91.03} & \textbf{67.85} & \textbf{85.00} & \textbf{92.68} & \textbf{64.56} \\ \bottomrule
\end{tabular}}
\caption{\textbf{Ablation study on the impact of process quantity in Multi-Scale Reasoning on CIRR val dataset.} All experiments are performed with the ViT-B/32 CLIP model.}
\label{appdix_process_cirr}
\vspace{-0.15in}
\end{table*}

\section{Example of User Interventions}
\label{sec_appendix_user_interv}

CIRCoT enables a highly transparent and interpretable reasoning process, which facilitates error tracking and correction through user intervention when necessary. We demonstrate this capability through illustrative cases in Figure~\ref{appendix_user_interv}, where initial reasoning processes led to suboptimal retrieval results. The structured nature of our reasoning framework allows users to precisely identify problematic reasoning steps and initiate corrective interactions with the LVLM. As illustrated in Figure~\ref{appendix_user_interv}, we present instances where users successfully identified and addressed two reasoning errors: \textit{the bird is to the right of the dog}" and \textit{red surface}". Through subsequent dialogue-based refinement, the model's retrieval accuracy was effectively improved, highlighting the practical value of our interpretable reasoning approach.

\vspace{-0.1in}
\section{Ablation Study on Multi-Scale Reasoning}


Tables~\ref{appdix_process} and~\ref{appdix_process_cirr} present a comparative analysis of single-process versus dual-process approaches in the multi-scale reasoning module, evaluated on the Fashion-IQ and CIRR datasets using the ViT-B/32 CLIP model. ``One process" refers to generating all three responses with LVLM simultaneously in a single inference pass. ``Two processes" represents our default methodology, which conducts reasoning separately at different scales through independent inference processes. Analysis of Table~\ref{appdix_process} reveals that employing a single process not only compromises the effectiveness of image-scale reasoning but also diminishes the performance gains typically achieved through object-scale reasoning integration. In the more challenging CIRR dataset, as shown in Table~\ref{appdix_process_cirr}, while concurrent reasoning of target image caption and key objects enhances image-scale reasoning accuracy, the incorporation of object-scale reasoning results yields a marginal performance degradation. We attribute these observations to two primary factors: \textbf{(1)} The utilization of identical reasoning logic across both scales potentially limits the semantic richness of the reasoning outcomes. \textbf{(2)} Qwen2-VL's current capabilities in managing multiple concurrent tasks may be insufficient, where the increased cognitive load adversely affects the precision of the results.

\begin{figure}[t!]
\centering  
\includegraphics[width=1.0\linewidth]{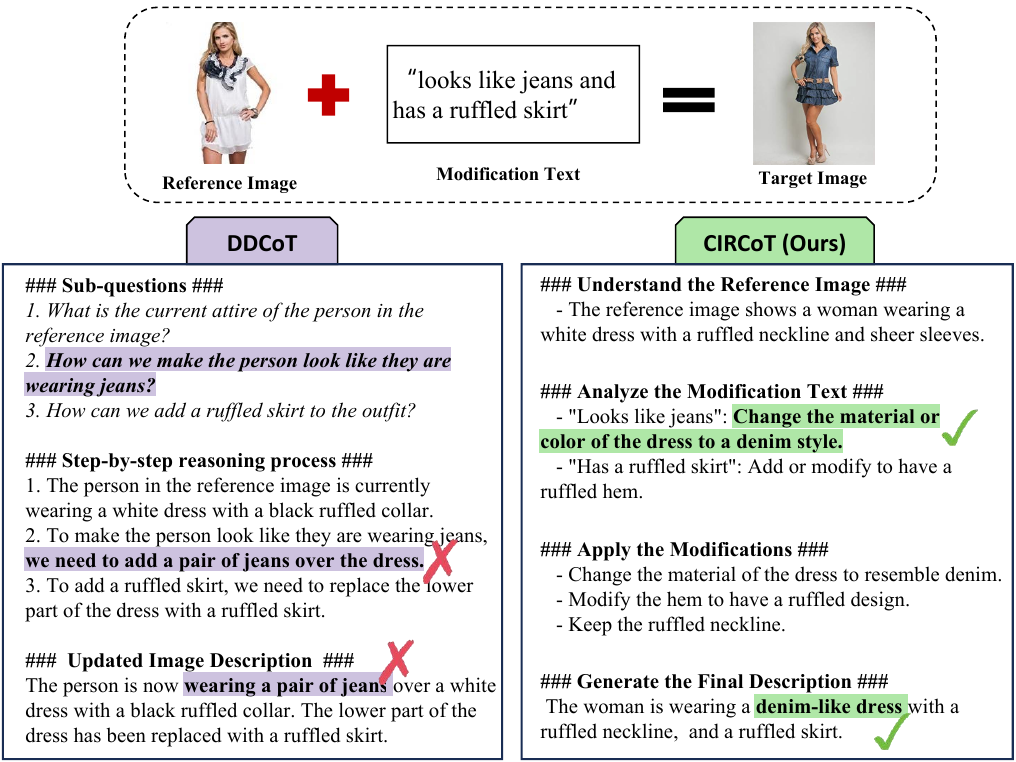} 
\caption{\textbf{Comparison between DDCoT and CIRCoT prompting strategies.}}
\label{appendix_ddcot_vs_circot}  
\vspace{-0.15in}
\end{figure}

\begin{table*}[]
\centering
\scalebox{0.85}{
\tabcolsep5.2pt
\begin{tabular}{clc|cccc||ccccccc|c}
\toprule
\multicolumn{3}{c|}{Benchmark} & \multicolumn{4}{c||}{\textbf{CIRCO}} & \multicolumn{8}{c}{\textbf{CIRR}} \\ \midrule
\multicolumn{3}{c|}{Metric} & \multicolumn{4}{c||}{mAP@k} & \multicolumn{4}{c|}{Recall@k} & \multicolumn{3}{c|}{Recall$_{sub}$@k} & \multirow{2}{*}{Avg.} \\
\multicolumn{1}{l}{Backbone} & Method & Training-free & k=5 & k=10 & k=25 & k=50 & k=1 & k=5 & k=10 & \multicolumn{1}{c|}{k=50} & k=1 & k=2 & k=3 &  \\ \midrule
\multirow{2}{*}{ViT-L/14} & LinCIR & \ding{55} & 12.62 & 13.40 & 14.81 & 15.69 & 25.08 & 53.63 & 67.30 & \multicolumn{1}{c|}{88.72} & 56.36 & 76.96 & 88.57 & 54.99  \\
 & \textbf{CoTMR} & \usym{1F5F8} & \textbf{27.61} & \textbf{28.22} & \textbf{30.61} & \textbf{31.70} & \textbf{35.02} & \textbf{64.75} & \textbf{76.18} & \multicolumn{1}{c|}{\textbf{92.51}} & \textbf{69.39} & \textbf{85.75} & \textbf{93.33} & \textbf{67.07} \\ \midrule
\multirow{2}{*}{ViT-G/14} & LinCIR & \ding{55} & 19.71 & 20.79 &	22.99 &	24.00 &	35.34 &	65.08 &	76.28 &	\multicolumn{1}{c|}{93.22} &	63.73 &	82.62 &	92.12 &	64.41  \\
 & \textbf{CoTMR} & \usym{1F5F8} & \textbf{31.73} & \textbf{32.72} & \textbf{35.30} & \textbf{36.43} & \textbf{36.36} & \textbf{66.92} & \textbf{77.82} & \multicolumn{1}{c|}{\textbf{93.99}} & \textbf{70.69} & \textbf{86.34} & \textbf{93.87} & \textbf{68.81} \\ \bottomrule
\end{tabular}}
\vspace{-0.08in}
\caption{\textbf{Comparison  with the LinCIR~\cite{lincir} on CIRCO and CIRR test sets.} Avg. indicates the average results of Recall@5 and Recall$_{sub}$@1. We reproduce the results of LinCIR. The best results are in boldface, while the second-best results are underlined.}
\label{appendix_tab_cirr_circo_main}
\vspace{-0.08in}

\end{table*}

\begin{table*}[t!]
\centering
\scalebox{0.8}{
\tabcolsep8pt
\begin{tabular}{llcccccccccc}
\toprule
\multirow{2}{*}{Backbone} & \multirow{2}{*}{Method} & \multirow{2}{*}{Training-free} & \multicolumn{2}{c}{Shirt} & \multicolumn{2}{c}{Dress} & \multicolumn{2}{c}{Tops\&Tee} & \multicolumn{3}{c}{Avg.} \\ \cline{4-12} 
 & &  & R@10 & R@50 & R@10 & R@50 & R@10 & R@50 & R@10 & R@50 & R$_{mean}$ \\ \hline
\multirow{2}{*}{ViT-L/14} & LinCIR & \ding{55} & 28.81 & 46.61 & 20.72 & 41.7 & 29.07 & 49.77 & 26.21 & 46.03 & 36.12 \\
 & \textbf{CoTMR} & \usym{1F5F8} & \textbf{35.43} & \textbf{54.91} & \textbf{31.18} & \textbf{55.04} & \textbf{38.55} & \textbf{61.33} & \textbf{35.05} & \textbf{57.09} & \textbf{46.50} \\ \hline
\multirow{3}{*}{ViT-G/14} & LinCIR & \ding{55} & {\ul 44.08} & {\ul 62.56} & {\ul 38.48} & {\ul 60.63} & {\ul 48.58} & \textbf{69.10} & {\ul 43.71} & {\ul 64.10} & {\ul 53.91}
 \\
& CoTMR & \usym{1F5F8} & {38.32} & {62.24} & {33.96} & {56.22} & {40.90} & {64.30} & {37.72} & {60.92} & {49.32} \\
 & \textbf{CoTMR + LinCIR} & \ding{55} & \textbf{46.24} & \textbf{67.95} & \textbf{40.06} & \textbf{63.11} & \textbf{49.59} & {\ul 68.03} & \textbf{45.13} & \textbf{66.36} & \textbf{55.74} \\ \bottomrule
\end{tabular}}
\vspace{-0.08in}
\caption{\textbf{Comparison with LinCIR~\cite{lincir} on the Fashion-IQ dataset.} R$_{mean}$ indicates the average results across all the metrics. We reproduce the results of LinCIR. The best results are in boldface, while the second-best results are underlined.}
\label{appendix_tab_fashioniq_main}
\vspace{-0.15in}
\end{table*}

\section{Qualitative Comparison of DDCoT and CIRCoT}
In Figure~\ref{appendix_ddcot_vs_circot}, we compare the reasoning processes of DDCoT and CIRCoT. It can be observed that when using DDCoT, the definition of subtasks is entirely left to the LVLM, which sometimes results in confusing subtasks, such as ``\textit{How can we make the person look like they are wearing jeans?}" in Figure~\ref{appendix_ddcot_vs_circot}. Such subtasks can mislead the LVLM into providing incorrect answers, as seen in Figure~\ref{appendix_ddcot_vs_circot}, where the LVLM decides on ``\textit{a pair of jeans}" to correspond to the modification text's requirement of ``\textit{looks like jeans}". This example highlights the critical importance of subtask definition; inappropriate subtasks can lead to erroneous logical reasoning and cause confusion. In contrast, our proposed CIRCoT, with its four predefined subtasks, offers a stable and correct reasoning process, leading to more accurate outputs. 

\begin{figure*}[t!]
\centering  
\includegraphics[width=1.0\linewidth]{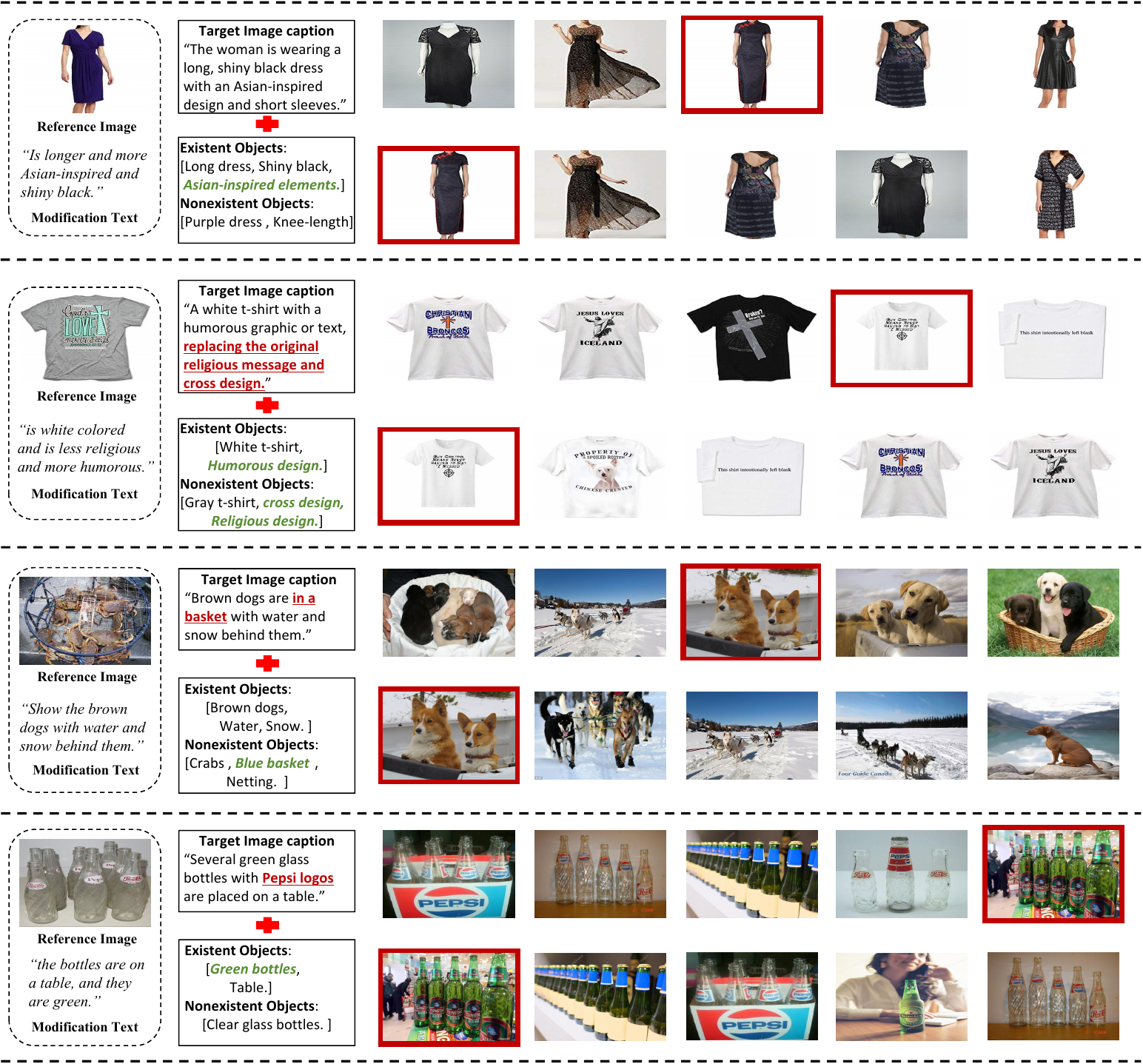} 
\caption{\textbf{Successful retrieval examples with muti-scale reasoning from Fashion-IQ and CIRR val set.} The ground-truth image is highlighted with the red box. Red underlined text indicates distracting information that causes mistake retrieval, while green italicized text represents key objects that help in correct retrieval.}
\label{appendix_qua_retrieval}  
\vspace{-0.1in}
\end{figure*}

\section{More Qualitative Examples}
\label{app_sec:Qualitative}
Figure~\ref{appendix_qua_retrieval} visualizes more cases where the combination of image-scale and object-scale reasoning leads to successful retrievals on both Fashion-IQ and CIRR datasets. \textbf{(1) }In the first example, the top two images retrieved using the target image caption overlooked the semantics of ``\textit{Asian-inspired design}". However, after emphasizing this part with existent objects, the target image was successfully ranked first. \textbf{(2)} In the second example, the target image caption contained distracting information (religious message and cross design), causing the top three images to include some religious elements. By incorporating nonexistent objects, our model successfully reduced the impact of this distracting information. \textbf{(3)} In the third example, the reference image had very little relevance to the target image, meaning the model could easily be misled by distracting information from the reference image, such as ``\textit{in a basket}". Object-scale reasoning can reduce such distractions because it doesn't need to consider the logical relationships between objects. Thus, our model successfully categorized ``\textit{Blue basket}" as a nonexistent object. \textbf{(4)} In the fourth example, the target image caption was similarly affected by the ``\textit{Pepsi logos}" in the reference image. Our model mitigated the distracting influence by emphasizing ``\textit{Green bottles}" and successfully ranked the target image first.

\section{Comparison and Integration with LinCIR}
Currently, there are two mainstream approaches in ZS-CIR: pseudo-token-based methods~\cite{saito2023pic2word, baldrati2023zero, lincir} and textual caption-based methods (including LLM-based~\cite{CIReVL, ldre, GRB} methods and LVLM-based methods, such as our CoTMR). In this section, we compare CoTMR with LinCIR~\cite{lincir}, a state-of-the-art pseudo-token-based method, and preliminarily explore the potential for their collaboration. 

Based on the results shown in Tables~\ref{appendix_tab_fashioniq_main} and \ref{appendix_tab_cirr_circo_main}, we observe that: \textbf{(1)} With ViT-L/14 CLIP, CoTMR achieves significant performance gains over LinCIR across all metrics on three datasets, demonstrating remarkable improvements such as a \textbf{12.08\%} increase in Avg. on CIRR and a \textbf{10.38\%} increase in R$_{mean}$ on Fashion-IQ. These results strongly validate the effectiveness of CoTMR and suggest that textual descriptions are more compatible with smaller-scale CLIP models compared to pseudo-tokens. \textbf{(2)} When employing ViT-G/14 CLIP, CoTMR significantly outperforms LinCIR on both CIRCO and CIRR datasets, achieving notable improvements such as a \textbf{12.02\%} increase in mAP@5 on CIRCO. However, we observe that LinCIR shows considerable advantages on the Fashion-IQ dataset. This observation indicates that pseudo-token-based methods may be more sensitive to specific CLIP versions or training datasets, while our approach demonstrates more consistent improvements across different CLIP versions.

Furthermore, we conduct a preliminary investigation into the potential of combining pseudo-token-based and textual caption-based methods to achieve superior performance. Specifically,  we first convert the reference image into a pseudo-token following LinCIR. Then, for the multi-scale reasoning outputs in CoTMR, we concatenated the pseudo-token with both the ``\textit{target image caption}" and the combination of ``\textit{existent objects}", while maintaining the ``\textit{nonexistent objects}" unchanged. We then retrieved target images using the same scoring mechanism as CoTMR. As shown in the last row of Table~\ref{appendix_tab_fashioniq_main}, CoTMR with additional pseudo-tokens achieves substantial improvements (e.g., a \textbf{6.42\%} increase in R$_{mean}$). Similarly, the multi-grained descriptions generated through CoTMR's reasoning process help enhance LinCIR's performance (e.g., a \textbf{1.83\%} improvement in R$_{mean}$). This experiment suggests a promising direction for ZS-CIR research: optimizing the text that concatenated with pseudo-tokens with LVLM. We leave this exploration for future work.


\end{document}